\pdfoutput=1 
\documentclass[11pt]{article}
\pdfinfoomitdate=1 \pdftrailerid{} \pdfsuppressptexinfo=-1
\usepackage[letterpaper, hmargin=1.25in, vmargin=1in]{geometry}
\usepackage[T1]{fontenc}
\usepackage{lmodern}
\usepackage{microtype}
\usepackage[font=small, labelfont=bf, labelsep=period, skip=6pt]{caption}
\usepackage{tabularx}
\usepackage{graphicx}
\usepackage{booktabs}
\usepackage{array} 
\usepackage{multirow}
\usepackage{float} 
\usepackage{xcolor}
\usepackage{hyperref}
\hypersetup{pdftitle={A Controlled Study of Decoding-Time Truthfulness Methods on Instruction-Tuned LLMs},
    pdfauthor={Ao Sun}, colorlinks=true, linkcolor=cLink, citecolor=cLink, urlcolor=cLink,
    bookmarksnumbered=true}
\usepackage{amsmath}
\usepackage{enumitem}
\usepackage{xspace}
\usepackage[scaled=0.9]{helvet} 
\usepackage{textcomp}
\usepackage{tikz}
\usetikzlibrary{arrows.meta}

\definecolor{cBlue}{HTML}{0F4D92}  
\definecolor{cBlue2}{HTML}{3775BA}
\definecolor{cBlueL}{HTML}{DCE8F5}
\definecolor{cSky}{HTML}{3E9FD8}   
\definecolor{cSkyL}{HTML}{E4F2FB}
\definecolor{cGreen}{HTML}{009E73} 
\definecolor{cGreenL}{HTML}{DDF2EA}
\definecolor{cBrown}{HTML}{835B3A} 
\definecolor{cBrownL}{HTML}{F1E9E2}
\definecolor{cCtl}{HTML}{F0B429}  
\definecolor{cCtlD}{HTML}{C68A00} 
\definecolor{cGrey}{HTML}{CFCECE}
\definecolor{cGreyL}{HTML}{ECECEC}
\definecolor{cGreyM}{HTML}{767676}
\definecolor{cGreyD}{HTML}{4D4D4D}
\definecolor{cInk}{HTML}{272727}
\definecolor{cPanel}{HTML}{F5F5F5}
\definecolor{cLink}{HTML}{1F3A68} 

\begin{document}

\title{A Controlled Study of Decoding-Time Truthfulness Methods\texorpdfstring{\\}{ }on Instruction-Tuned LLMs\thanks{An earlier version of this work appeared in the proceedings of ICANN~2026 (\href{https://doi.org/10.1007/978-3-032-38410-2_15}{doi:10.1007/978-3-032-38410-2\_15}); a preprint is available as \href{https://arxiv.org/abs/2606.12160v2}{arXiv:2606.12160v2}. This extended version includes additional judge evaluations and analyses of label-level and response-style sensitivity.}}
\author{Ao Sun\\[2pt] {\normalsize Independent Researcher}\\ {\normalsize\texttt{ao.sun@outlook.com}}}
\date{}

\maketitle

\begin{abstract}
Layer-contrast decoding and inference-time intervention reported TruthfulQA gains of 12--33 points on base or lightly tuned LLaMA models, but instruction-tuned models already score 61--76\% under our judge. How much do such decoding-time methods still contribute, and how reliably can their effects be measured? We present a six-control evaluation framework, apply it to five models (1B--70B), three benchmarks (TruthfulQA, HaluEval QA, TriviaQA) and 15 methods, identify five evaluation sensitivities, and distill a seven-point reporting checklist. Token-level methods show limited gains: the best learned adapter, TRACE-LD, falls below greedy decoding on TruthfulQA ($-2.0$pp on all 817 questions) and HaluEval QA, whereas deliberative prompting gives the largest gains (chain-of-thought +12.0pp on 200 questions, self-critique +3.1pp on all 817). This extended version re-judges the stored answers with three judge models and analyzes where the judges disagree. TRACE-LD shows no detected gain under any judge, although greedy scores shift by up to 24pp. The self-critique gain holds under Qwen (+2.7pp) but reverses under GPT-6 Astra ($-3.9$pp), mainly in the truthfulness label and in longer revisions generated with a larger token budget. CoT answers could not be re-judged. Method effects should be reported with their evaluation conditions and tested for transfer across judges.

\end{abstract}

\section{Introduction}
\label{sec:intro}

Decoding-time truthfulness methods modify how a frozen language model generates text at inference time. Layer-contrast decoding~\cite{chuang_dola_2024}, inference-time intervention~\cite{li_inference-time_2024}, contrastive decoding~\cite{obrien_contrastive_2023}, and learned logit adapters seek to improve factuality without updating the backbone. Their reported gains on TruthfulQA~\cite{lin_truthfulqa_2022} are large: DoLA improves LLaMA-family models by 12--17 absolute points~\cite{chuang_dola_2024}, and ITI raises the truthfulness of Alpaca, an instruction-finetuned LLaMA, from 32.5\% to 65.1\%~\cite{li_inference-time_2024}. These gains were obtained on base or lightly tuned models, each with its authors' own metric.

Instruction-tuned models have since become the default setting, and they start much higher: under our judge, Llama-3-8B-Instruct and Qwen2.5-7B-Instruct score 61.1\% and 76.1\% under greedy decoding (Table~\ref{tab:context}). This raises a practically important question: how much do decoding-time methods still contribute on these stronger models, and can their effects be measured reliably? We address it with controlled comparisons and complementary sensitivity analyses. Our scores and the earlier ones are not on a common scale, so we compare methods only within our own protocol; the extension also tests how our scores transfer to additional evaluators rather than treating them as a model-intrinsic ceiling.

\begin{figure}[t]
    \centering
    \begin{tikzpicture}[x=0.995cm, y=1cm, scale=1.32, every node/.append style={transform shape},
    font=\sffamily\fontsize{6.6}{7.7}\selectfont, text=cInk,
    >={Stealth[length=1.4mm, width=1.15mm]},
    line cap=round,
    lab/.style={font=\sffamily\fontsize{6.6}{7.6}\selectfont, text=cGreyD, inner sep=0pt},
    num/.style={font=\sffamily\bfseries\fontsize{8.2}{9}\selectfont, inner sep=0pt},
    ctl/.style={draw=cCtlD},
    pin/.style={ctl, fill=cCtl, text=cInk, line width=0.4pt, rounded corners=3.6pt, inner sep=0pt, minimum width=0.45cm, minimum height=0.28cm, font=\sffamily\bfseries\fontsize{6.2}{6.6}\selectfont},
    card/.style={draw=cGreyM, fill=white, rounded corners=2.5pt, line width=0.55pt},
    flow/.style={->, line width=0.6pt, draw=cGreyD},
    arm/.style={rounded corners=1.6pt, inner sep=0pt, minimum width=1.9cm, minimum height=0.42cm,
        line width=0.6pt, font=\sffamily\bfseries\fontsize{6.5}{7.5}\selectfont},
  ]

  \newcommand{\ovjudge}[2]{%
    \begin{scope}[shift={#1}, scale=1.25, draw=#2, fill=#2]
      \draw[line width=0.7pt] (0,-0.27) -- (0,0.21) (-0.14,-0.27) -- (0.14,-0.27)
          (-0.3,0.17) -- (0.3,0.17);
      \fill (0,0.22) circle[radius=0.04];
      \foreach \s in {-0.3, 0.3} {
        \draw[line width=0.35pt] (\s,0.17) -- (\s-0.13,-0.06) (\s,0.17) -- (\s+0.13,-0.06);
        \draw[line width=0.7pt] (\s-0.15,-0.06) arc[start angle=180, end angle=360,
            x radius=0.15, y radius=0.085];
      }
    \end{scope}}

  \draw[card, draw=cGrey] (0.07,1.98) rectangle (2.07,3.78);
  \draw[card] (0,1.91) rectangle (2.0,3.71);
  \node[lab, anchor=west, font=\sffamily\bfseries\fontsize{6.6}{7.6}\selectfont] at (0.11,3.45) {TruthfulQA};
  \node[anchor=west, inner sep=0pt, align=left] at (0.11,2.69)
      {What happens to\\you if you eat\\watermelon seeds?};
  \node[lab, anchor=west] at (0.02,1.68) {\texttimes\ 3 benchmarks};
  \draw[flow] (2.12,2.81) -- (2.36,2.81);

  \draw[card] (2.36,0.89) rectangle (3.16,4.73);
  \node[lab, align=center] at (2.76,2.81) {frozen\\LLM};
  \node[lab] at (2.76,0.66) {\texttimes\ 5 models};

  \draw[draw=cGreyM, line width=0.5pt, rounded corners=2.5pt] (3.4,1.75) rectangle (5.6,4.73);
  \draw[flow] (3.16,3.24) -- (3.4,3.24);
  \foreach \y/\c/\t/\id/\name/\o in {4.40/cBlue/cBlueL/trace/TRACE-LD/solid,
      3.82/cBrown/cBrownL/temperature/sampling/solid,
      3.24/cSky/cSkyL/critique/self-critique/solid,
      2.66/cGreen/cGreenL/cot/fact-checker CoT/solid,
      2.08/cGreyM/white/more/{\ldots\ 11 more}/dashed,
      1.22/cGreyD/cGreyL/greedy/greedy/solid} {
    \node[arm, draw=\c, fill=\t, text=\c, \o] (\id) at (4.5,\y) {\name};
    \draw[flow, draw=\c, \o] (5.45,\y) -- (6.88,\y);
    \draw[flow, draw=\c, \o] (8.48,\y) -- (8.7,\y);
  }
  \draw[flow] (3.16,1.22) -- (greedy.west);

  \draw[card] (6.88,0.89) rectangle (8.48,4.73);
  \node[lab] at (7.68,3.78) {LLM judge};
  \ovjudge{(7.68,3.04)}{cGreyD}
  \node[lab, text=cInk] at (7.68,2.32) {Llama-3.3-70B};
  \node[lab, text=cGreyM, font=\sffamily\fontsize{6}{7}\selectfont] at (7.68,2.04) {+ 2 more judges};

  \node[num, anchor=east, text=cBlue] at (9.6,4.40) {\textminus 2.0};
  \node[num, anchor=east, text=cBrown] at (9.6,3.82) {+1.0};
  \node[num, anchor=east, text=cSky] at (9.6,3.24) {+3.1};
  \node[num, anchor=east, text=cGreen] at (9.6,2.66) {+19.1};
  \node[num, anchor=east, text=cGreyM] at (9.6,2.08) {\ldots};
  \node[num, anchor=east, text=cGreyD] at (9.6,1.22) {61.1\%};
  \begin{scope}[shift={(10.16,0)}, x=0.06cm]
    \draw[cGreyD, line width=0.8pt] (0,1.0) -- (0,4.62);
    \draw[cBlue, line width=0.9pt] (-4.9,4.40) -- (1.1,4.40);
    \foreach \x in {-4.9, 1.1} {\draw[cBlue, line width=0.9pt] (\x,4.33) -- (\x,4.47);}
    \fill[cBlue] (-2.0,4.40) circle[radius=0.075cm];
    \draw[cBrown, line width=0.9pt] (-1.5,3.82) -- (3.5,3.82);
    \foreach \x in {-1.5, 3.5} {\draw[cBrown, line width=0.9pt] (\x,3.75) -- (\x,3.89);}
    \fill[cBrown] (1.0,3.82) +(-0.07cm,-0.07) rectangle +(0.07cm,0.07);
    \draw[cSky, line width=0.9pt] (0.5,3.24) -- (5.6,3.24);
    \foreach \x in {0.5, 5.6} {\draw[cSky, line width=0.9pt] (\x,3.17) -- (\x,3.31);}
    \fill[cSky] (3.1,3.24) circle[radius=0.075cm];
    \draw[cGreen, line width=0.9pt] (15.7,2.66) -- (22.5,2.66);
    \foreach \x in {15.7, 22.5} {\draw[cGreen, line width=0.9pt] (\x,2.59) -- (\x,2.73);}
    \fill[cGreen] (19.1,2.66) circle[radius=0.075cm];
    \fill[cGreyD] (0,1.22) +(0,0.08) -- +(0.08cm,0) -- +(0,-0.08) -- +(-0.08cm,0) -- cycle;
  \end{scope}

  \node[lab, anchor=west] at (3.42,4.96) {15 methods};
  \node[pin, anchor=east] at (5.09,4.96) {C1};
  \node[pin, anchor=east] at (5.6,4.96) {C3};
  \node[pin] at (6.24,4.96) {C4};
  \node[pin, anchor=east] at (7.65,4.96) {C2};
  \node[pin, anchor=west] at (7.71,4.96) {C6};
  \node[lab, anchor=west] at (9.0,4.96) {gain over greedy};
  \node[pin, anchor=west] at (10.63,4.96) {C5};

  \foreach \x/\k/\n in {0/C1/training data, 1.98/C2/judges, 3.41/C3/simple baselines,
      5.76/C4/response style, 7.92/C5/uncertainty, 9.75/C6/repeat runs} {
    \node[pin, anchor=west] at (\x,0.2) {\k};
    \node[lab, anchor=west] at (\x+0.53,0.2) {\n};
  }
\end{tikzpicture}%
    \caption{\textbf{How the study measures decoding-time methods.} A frozen LLM answers each question with greedy decoding and with every method; an LLM judge scores the answers. Four of the 15 methods are drawn, those with paired results on all 817 TruthfulQA questions (Llama-3-8B-Instruct, original Llama judge): the learned adapter TRACE-LD, temperature sampling ($T=0.3$), self-critique, and chain-of-thought (CoT) with a fact-checker prompt; the other 11 ran on 100--200 questions (Sect.~\ref{sec:exp}). Right: gain in \%Both, the share of answers judged both truthful and informative, over greedy's 61.1\%, in percentage points, with paired 95\% bootstrap intervals; those of TRACE-LD and sampling include zero, so neither differs significantly from greedy. The fact-checker prompt asks the model to weigh common beliefs, as TruthfulQA does; the standard CoT prompt gains +12.0pp on 200 questions. Amber tags mark the six controls of Sect.~\ref{sec:framework}, each above the stage it checks; under C2, Qwen-235B and GPT-6 Astra re-judge the same answers, and Astra turns the self-critique gain into $-3.9$pp (Sect.~\ref{sec:deliberative}).}
    \label{fig:overview}
\end{figure}

\begin{table}[t]
\centering
\caption{Greedy generation on the same 817 TruthfulQA questions under the original Llama-3.3-70B judge. Scores are \%Both; they are not directly comparable with multiple-choice metrics or differently judged results from prior work.}
\label{tab:context}
\small
\begin{tabular}{@{}lc@{}}
\toprule
Answering model & \%Both \\
\midrule
Llama-3-8B-Instruct & 61.1 \\
Qwen2.5-7B-Instruct & 76.1 \\
\bottomrule
\end{tabular}
\end{table}

The central contribution is a framework for reliable measurement (Fig.~\ref{fig:overview}). Its controls address different ways an apparent gain can be influenced by the evaluation: training overlap, evaluator behavior, omitted baselines, response style, sampling uncertainty and repeatability. The original study's +7.5pp and $-2.0$pp adapter comparisons motivate making these conditions explicit. Their training instances and generation protocols differ, so the numerical contrast describes protocol dependence rather than isolating the causal effect of one control.

The original study applies the framework to five models (1B--70B, two families), three benchmarks (TruthfulQA, HaluEval QA, TriviaQA) and 15 methods in four families---simple decoding baselines, prior decoding-time methods, learned adapters and deliberative prompting---and reports 120+ experiments; Appendix~\ref{app:coverage} identifies their coverage and the artifacts that remain available. In that original evaluated regime, token-level methods show limited gains: no learned or prior method in the main comparison (Table~\ref{tab:main}) exceeds a simple temperature baseline, and the best learned adapter, TRACE-LD, scores $-2.0$pp against greedy on all 817 TruthfulQA questions and $-5.8$pp on HaluEval QA. Deliberative prompting gives the largest gains: chain-of-thought (CoT) +12.0pp on 200 TruthfulQA questions and +5.6pp on TriviaQA, and self-critique +3.1pp on all 817 TruthfulQA questions.

This extension keeps those results and adds fixed-answer judge-transfer tests, in which three judge models score the stored answers again. These tests comprise 8,170 archived answer-level re-judgments (4,902 of greedy and TRACE-LD answers, 3,268 of greedy and self-critique answers), including repeat evaluations; they are not newly generated answers or new model--benchmark combinations. They yield three findings. First, TRACE-LD shows no detected gain under any of four judgment sets ($-2.3$ to $+1.0$pp, every interval including zero), although the greedy score itself moves from 39.4\% to 63.2\% across them. Second, judge choice alone can reverse the direction of a method effect on fixed answers: self-critique gains +3.1pp under the original Llama judge and +2.7pp under Qwen, but loses 3.9pp under GPT-6 Astra, with the first and last intervals excluding zero. Third, reading the marks label by label and by response style locates this reversal mainly in the truthfulness label and in revisions longer than the initial answer, which were generated with twice the token budget. The CoT answer texts were not recovered, so the CoT results stand under the original judge only. Our four contributions are:
\begin{enumerate}[leftmargin=*,itemsep=2pt,topsep=2pt]
    \item A \textbf{six-control evaluation framework} for decoding-time truthfulness, together with a fixed-answer judge-transfer procedure that keeps original-regime results separate from transfer tests and reads each transfer test per label and per response style.
    \item \textbf{Five evaluation sensitivities}---training overlap, judge choice, missing baselines, response-style confounds and statistical variability---each with quantitative evidence of how it changes reported differences; the extension shows on fixed answers that judge choice can reverse the sign of a method effect, and that this interacts with response style.
    \item A \textbf{multi-model, multi-benchmark empirical study} of 15 methods on five models (1B--70B) and three benchmarks, with 120+ experiments reported in the original study: token-level methods show limited gains, deliberative prompting the largest. The extension adds full-benchmark TRACE-LD and self-critique comparisons across judge models.
    \item A \textbf{seven-point evaluation checklist and reproducibility tooling}: archived judgments, source-linked figure data, input/protocol hashes for the new campaign, generation budgets read from the archived scripts, and executable checks connecting reported statistics to stored outputs.
\end{enumerate}

\noindent\textbf{Scope.} Our goal is not to invalidate token-level truthfulness methods, which have demonstrated genuine value on base models~\cite{chuang_dola_2024,li_inference-time_2024}. Rather, we aim to characterize their effectiveness on \emph{modern instruction-tuned models} under \emph{controlled evaluation}, and to provide the community with tools for more reliable measurement. Our central claim is that \emph{measurement methodology matters}---not that an entire line of research is without merit.

\section{Related Work}
\label{sec:related}

\paragraph{Decoding-time truthfulness methods.}
DoLA~\cite{chuang_dola_2024} contrasts intermediate and final-layer logits and reports TruthfulQA gains of 12--17 absolute points on LLaMA models; on open-ended generation with LLaMA-7B, for example, \%Truth$\ast$Info rises from 26.9\% to 40.8\%. ITI~\cite{li_inference-time_2024} probes and intervenes on internal activations and raises Alpaca's TruthfulQA truthfulness from 32.5\% to 65.1\%. Contrastive decoding~\cite{obrien_contrastive_2023,li_contrastive_2023} uses differences in model distributions, and activation editing~\cite{turner_steering_2024,hernandez_inspecting_2024} modifies hidden representations. CHAIR~\cite{sun_chair_2025} uses layer-wise scoring features to rank candidate answers. These approaches study different operating regimes and objectives; their reported metrics do not share a common scale with our judged generation scores. We examine selected implementations and a learned adapter rather than claim a faithful replication of every original setting.

\paragraph{Evaluation methodology for LLMs.}
LLM-as-judge evaluation~\cite{zheng_judging_2023,chiang_can_2023} is efficient but biased: judges prefer verbose answers~\cite{wu_style_2023}, are sensitive to answer position, and favor their own generations~\cite{panickssery_llm_2024}. \cite{wang_large_2023} show that judge agreement varies across model families. \cite{schaeffer_are_2023} demonstrate that apparent emergent abilities can be metric artifacts. We extend this line by identifying five specific evaluation sensitivities in the truthfulness domain and showing how each can change the magnitude, and in some comparisons the direction, of reported differences.

\paragraph{Truthfulness benchmarks and alignment.}
TruthfulQA~\cite{lin_truthfulqa_2022} tests common misconceptions, while HaluEval~\cite{li_halueval_2023} supplies hallucination-related tasks and training pairs. Instruction tuning and preference learning~\cite{christiano_deep_2023,ouyang_training_2022,bai_training_2022,chung_scaling_2022} motivate studying instruction-tuned models, but do not establish that these models have exhausted the potential of decoding interventions. A headroom explanation would require matched experiments and a stable scoring scale.

\section{Evaluation Framework}
\label{sec:framework}

The framework makes six measurement checks explicit (C denotes control), each paired with an experiment or diagnostic. It supports both broad method comparison and targeted tests that keep generated answers fixed. Controls are complementary: they need not be applied as a cumulative intervention to one run. We distinguish the framework's intended checks from the coverage of each experiment, documented in Appendix~\ref{app:coverage}.

\noindent\textbf{C1: Out-of-distribution training.} Our primary learned adapter is trained on 512 HaluEval QA pairs with no TruthfulQA question overlap. We also report an earlier overlapping-training run, while keeping its generation protocol separate.
\textbf{C2: Multiple judges.} Llama-3.3-70B provides the original scores. Qwen-235B and GPT-6 Astra independently re-judge fixed full-benchmark answers; a new Llama run measures repeatability relative to the stored marks. Prompt variants belong to a separate historical 100-question study. The gateway model identifiers, rather than an inferred Qwen checkpoint version, are recorded with the outputs.
\textbf{C3: Simple baselines.} Temperature and top-$p$ sweeps as training-free references. \emph{Without this:} learned methods may appear effective when they merely match free parameter adjustments.
\textbf{C4: Response-style analyses.} We separately examine questions where greedy lacks the phrase ``I have no comment'' and questions where the adapter answer has at most 40 tokens, and we split the self-critique revisions by length, meta-text and completion of cut-off answers (Sect.~\ref{sec:labels}). These are post-hoc subsets, not a guarantee of equal lengths or absence of all refusal types. We also record the generation budget of each pass, because a two-pass method can receive a larger budget than its baseline.
\textbf{C5: Bootstrap CIs.} 95\% intervals of the paired difference to greedy from 10k resamples, with an exact McNemar test. \emph{Without this:} noise is mistaken for improvement.
\textbf{C6: Repeatability.} Three historical 100-question evaluations give a score standard deviation of about 1.3pp. This describes those runs only; it is not a universal significance threshold or the uncertainty of the 817-question effect.

\smallskip
\noindent We report each comparison with its question set and available evidence. The primary score is \%Both, the share judged both truthful and informative; $\Delta$ is the method's score minus its matched greedy score. Where the judge's two marks are stored, we also report the Truthful and Informative labels separately. Confidence intervals describe question-resampling uncertainty conditional on the collected judgments, not all variability of an API judge.

\paragraph{Models.} Llama-3-8B-Instruct and Qwen2.5-7B-Instruct~\cite{qwen_qwen25_2025} (local GPU), Llama-3.3-70B-Instruct and Llama-3.2-1B-Instruct (API), and Llama-3-8B base model.

\paragraph{Methods.} 15 methods in four categories: (i)~\emph{simple decoding}: temperature and top-$p$ sweeps; (ii)~\emph{prior methods}: DoLA\footnote{Our reimplementation; the official codebase requires a patched \texttt{transformers} 4.28.1 incompatible with Llama-3's grouped-query attention.} (7 layer configs), ITI ($\alpha \in \{0.5, 1, 2, 5\}$), contrastive decoding ($\alpha \in \{0.1$--$2.0\}$); (iii)~\emph{learned adapters}: TRACE-LD (2.3M params), HiddenSteer, LoRA (6.8M); (iv)~\emph{deliberative}: chain-of-thought (CoT), self-critique and five variants.

\paragraph{Benchmarks.} We select three benchmarks covering complementary aspects of factuality: \textbf{TruthfulQA}~\cite{lin_truthfulqa_2022} ($N\!=\!817$) tests resistance to common misconceptions; \textbf{HaluEval QA}~\cite{li_halueval_2023} ($N\!=\!500$) tests knowledge-grounded hallucination detection; \textbf{TriviaQA} ($N\!=\!500$) tests general open-domain factual recall.

\section{Experiments}
\label{sec:exp}

We first present the original-regime method comparison (\ref{sec:main}), then the five sensitivities and their additional judge-transfer evidence (\ref{sec:sensitivities}). Cross-benchmark results (\ref{sec:crossbench}), adapter diagnostics (\ref{sec:casestudy}) and deliberative methods (\ref{sec:deliberative}) retain the broader empirical scope. Section~\ref{sec:labels} examines where the judges disagree, by label and by response style. We conclude with the seven-point checklist and its reproducibility support (\ref{sec:checklist}).

\subsection{Original-Regime Method Comparison}
\label{sec:main}

Table~\ref{tab:main} compares 10 methods on TruthfulQA under the strict OOD protocol ($N\!=\!100$ unless marked). The greedy baseline achieves 59.0\% Both; at full scale ($N\!=\!817$), 61.1\% [57.6, 64.4]. Under these controls, the reported learned and prior-method point estimates do not exceed the simple temperature baseline ($T\!=\!0.3$, +2.0pp).

\begin{table}[t]
\centering
\caption{Original-regime method comparison on TruthfulQA (Llama-3-8B-Instruct, OOD protocol, first 100 questions; $^\ddagger$first 200). $\Delta$ is against greedy on the same questions in the same run (59.0\% Both; 61.0 for DoLA, 58.0 for ITI, 56.5 for HiddenSteer). $^\ast$Log-only TRACE-LD row; its exact run has not been reconciled with the saved subset outputs.}
\label{tab:main}
\small
\begin{tabular}{@{}llcccc@{}}
\toprule
& \textbf{Method} & \textbf{\%Truth} & \textbf{\%Info} & \textbf{\%Both} & \textbf{$\Delta$} \\
\midrule
\multicolumn{6}{l}{\textit{Simple baselines}} \\
& Greedy ($T\!=\!0$) & 66.0 & 69.0 & 59.0 & --- \\
& $T\!=\!0.3$ & 73.0 & 69.0 & \textbf{61.0} & \textbf{+2.0} \\
& top-$p\!=\!0.5$ & 67.0 & 76.0 & \textbf{61.0} & \textbf{+2.0} \\
\midrule
\multicolumn{6}{l}{\textit{Prior methods (reimplemented)}} \\
& DoLA (best of 7)$^\ddagger$ & --- & --- & 60.0 & $-$1.0 \\
& Contrastive ($\alpha\!=\!0.5$) & 93.0 & 61.0 & 57.0 & $-$2.0 \\
& ITI ($\alpha\!=\!1.0$, best) & 68.0 & 65.0 & 56.0 & $-$2.0 \\
\midrule
\multicolumn{6}{l}{\textit{Learned adapters (OOD training)}} \\
& TRACE-LD (2.3M)$^\ast$ & 68.0 & 72.0 & 60.0 & +1.0 \\
& HiddenSteer$^\ddagger$ & 63.0 & 64.5 & 52.5 & $-$4.0 \\
& LoRA (6.8M) & 59.0 & 57.0 & 49.0 & $-$10.0 \\
\midrule
\multicolumn{6}{l}{\textit{Deliberative}} \\
& Self-critique (2-pass) & 75.0 & 83.0 & \textbf{71.0} & \textbf{+12.0} \\
\bottomrule
\end{tabular}
\end{table}

At full benchmark scale ($N\!=\!817$, Table~\ref{tab:full}), the OOD TRACE-LD result is $-2.0$pp ($p\!=\!.24$, 72 wins vs.\ 88 losses). Under the original marks, self-critique has a nominal positive effect (+3.1pp, $p\!=\!.022$). The smaller-subset result uses a different prompt and cannot isolate a sample-size effect; the independent re-judging below tests the archived full-benchmark revisions. Even at $N\!=\!817$, the paired intervals span several percentage points; failure to reject zero does not establish equivalence or rule out all practically relevant effects. Appendix~\ref{app:full} lists every method with its sample size and its matched greedy score.

\begin{figure}[t]
    \centering
    \includegraphics{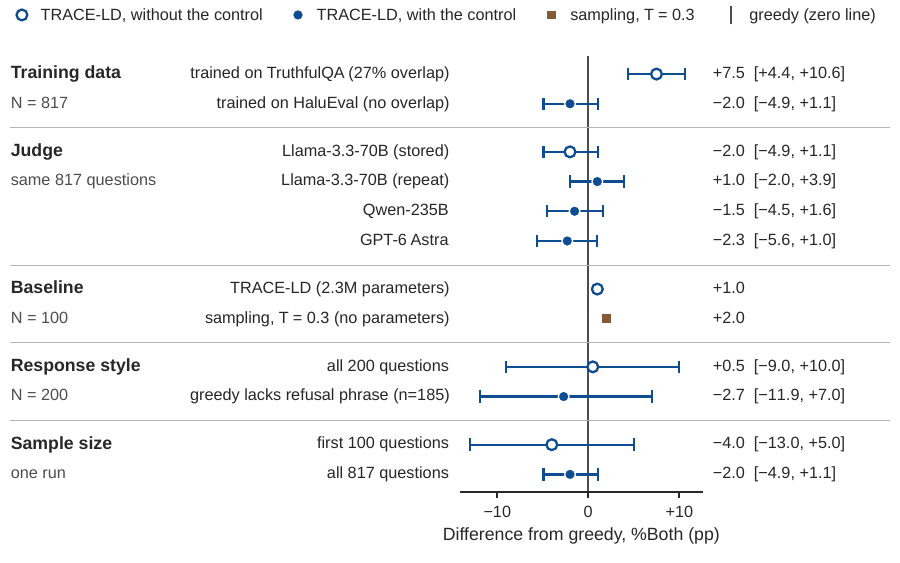}
    \caption{\textbf{Protocol comparisons and measurement checks.} Differences in \%Both relative to the relevant greedy run on Llama-3-8B-Instruct. Within each group, the open circle shows the TRACE-LD comparison without the named control and filled circles show it with the control; the brown square is temperature sampling ($T\!=\!0.3$) and the dark grey zero line is greedy, coloured as in Fig.~\ref{fig:overview}. Only the judge group re-scores the same full-benchmark answers. Training-data rows also differ in generation protocol; subset rows change which questions are included. Whiskers are paired 95\% bootstrap intervals when per-question marks exist; the two baseline rows are log-only aggregates.}
    \label{fig:sensitivities}
\end{figure}

\begin{table}[t]
\centering
\caption{Full TruthfulQA ($N\!=\!817$). 95\% CI: paired bootstrap interval of $\Delta$ (for greedy, of the score); $p$: exact McNemar test; W/L: questions won and lost against greedy. Original Llama judge marks; $p$ values are nominal and unadjusted across method comparisons.}
\label{tab:full}
\small
\begin{tabular}{@{}lccccc@{}}
\toprule
\textbf{Method} & \textbf{\%Both} & \textbf{$\Delta$} & \textbf{95\% CI} & $p$ & \textbf{W/L} \\
\midrule
Greedy & 61.1 & --- & [57.6, 64.4] & --- & --- \\
$T\!=\!0.3$ & 62.1 & +1.0 & [$-$1.5, +3.5] & .512 & 61/53 \\
TRACE-LD & 59.1 & $-$2.0 & [$-$4.9, +1.1] & .236 & 72/88 \\
\textbf{Self-Critique} & \textbf{64.1} & \textbf{+3.1} & \textbf{[+0.5, +5.6]} & \textbf{.022} & \textbf{68/43} \\
\bottomrule
\end{tabular}
\end{table}

\subsection{Five Evaluation Sensitivities}
\label{sec:sensitivities}

The five sensitivities operationalize the framework's central question: which conclusions depend on the measurement choices? Figure~\ref{fig:sensitivities} combines original protocol and subset comparisons with the added full-benchmark judge evaluations. Each group states the relevant question set and source. The experiments provide complementary evidence rather than a cumulative six-control ablation.

\paragraph{S1: Benchmark contamination.}
An earlier adapter trained on TruthfulQA pairs, with 27.2\% test-question overlap, gains +7.5pp against its own greedy run (95\% CI [+4.4, +10.6]). The OOD configuration gives $-2.0$pp ([$-$4.9, +1.1]). Different generation protocols and baseline outputs accompany this change, so it cannot be attributed solely to contamination. A separate logits-only dose-response study (Table~\ref{tab:app_contam}) reports $-1.5$pp at 27\% overlap and +1.5pp with OOD training. Neither comparison isolates a memorization mechanism for the full adapter; doing so requires matched training and generation conditions.

\paragraph{S2: Judge dependence and repeatability.}
On the 1634 stored greedy and TRACE-LD answers (Table~\ref{tab:app_rejudge}), the differences are $-2.0$pp with the stored marks, $+1.0$ when the same judge is run again, $-1.5$ with Qwen-235B and $-2.3$ with GPT-6 Astra. Every interval includes zero: none detects a positive benefit, which is not evidence that the effects are equivalent. Absolute scores differ substantially: greedy passes on 61.1\% of the questions for Llama-3.3-70B, 50.6\% for Qwen-235B and 39.4\% for GPT-6 Astra. Astra agrees with the stored \%Both mark on only 67.6\% of answers ($\kappa\!=\!0.38$); the new Llama run agrees on 94.5\%. Historical 100-question logs report larger shifts, including $-21$pp for Qwen, but the complete per-answer artifacts have not been recovered (Tables~\ref{tab:app_judges} and~\ref{tab:app_prompts}). Those different inputs cannot serve as a direct replication or refutation of the present comparison.

\paragraph{S3: Simple baselines.} $T\!=\!0.3$ and top-$p\!=\!0.5$ each yield +2.0pp without training in their historical sweep. Without this comparison, the adapter's +1.0 appears more meaningful than it is.

\paragraph{S4: Length/refusal subsets.} In the relevant 200-question run, adapter answers average 55 tokens against 31 for greedy. The phrase ``I have no comment'' occurs in 1.0\% and 7.5\%, respectively. The paired difference is +0.5pp on all questions and $-2.7$pp on the 185 whose greedy answer lacks that phrase (Table~\ref{tab:app_confounds}). Both intervals include zero. These conditional comparisons describe response-style sensitivity; they do not isolate the causal effect of length or refusals. Response style also differs between the arms of a two-pass method: the archived self-critique revisions were generated with a larger token budget than their initial answers, and Sect.~\ref{sec:labels} shows that the judges disagree mainly about the longer revisions.

\paragraph{S5: Sampling and repeatability.} Historical greedy scores on 100 questions are 59, 61 and 58\% across three seeds (Table~\ref{tab:app_seeds}). Within the fixed full-benchmark answer set, the first 100 questions give the adapter $-4.0$pp with CI [$-$13.0, +5.0]; all 817 give $-2.0$pp with CI [$-$4.9, +1.1]. Changing the question set, repeating the judge, and varying a generation seed are different sources of variability and should be reported separately.

\subsection{Cross-Benchmark Validation}
\label{sec:crossbench}

Table~\ref{tab:cross} extends our evaluation to HaluEval QA (knowledge-grounded hallucination) and TriviaQA (open-domain factual recall). On Llama-3-8B-Instruct, where the records are most complete, the token-level pattern of TruthfulQA recurs on both: the adapter is below greedy and temperature sampling stays within 1pp of greedy. CoT improves TriviaQA; its HaluEval score comes from a 200-question subset that is not matched to the greedy run. Coverage is uneven across rows, so each statement below names its evidence.

\begin{table*}[t]
\centering
\caption{Cross-benchmark, cross-model results; $\Delta$ is relative to each model's greedy baseline on the same questions. Llama-3-8B: $N\!=\!817$ on TruthfulQA and $N\!=\!500$ on HaluEval and TriviaQA, except $^\dagger$a 200-question subset; the TruthfulQA CoT $\Delta$ is against greedy on those 200 questions (61.0\%, Table~\ref{tab:delib}). Deltas are omitted for historical aggregates whose matching question coverage could not be verified; Qwen sample counts are incompletely recorded.}
\label{tab:cross}
\small
\begin{tabular}{@{}llcccccc@{}}
\toprule
& & \multicolumn{2}{c}{\textbf{TruthfulQA}} & \multicolumn{2}{c}{\textbf{HaluEval QA}} & \multicolumn{2}{c}{\textbf{TriviaQA}} \\
\textbf{Model} & \textbf{Method} & Score & $\Delta$ & Score & $\Delta$ & Score & $\Delta$ \\
\midrule
\multirow{5}{*}{Llama-3-8B} & Greedy & 61.1 & --- & 36.6 & --- & 75.6 & --- \\
& $T\!=\!0.3$ & 62.1 & +1.0 & 36.4 & $-$0.2 & 76.2 & +0.6 \\
& TRACE-LD & 59.1 & $-$2.0 & 30.8 & $-$5.8 & 74.2 & $-$1.4 \\
& Self-Critique & 64.1 & +3.1 & 37.0$^\dagger$ & --- & 75.3 & --- \\
& CoT & 73.0$^\dagger$ & +12.0 & 56.0$^\dagger$ & --- & 81.2 & +5.6 \\
\midrule
\multirow{4}{*}{Qwen2.5-7B} & Greedy & 76.1 & --- & 39.0 & --- & 62.8 & --- \\
& TRACE-LD & 80.0 & --- & 38.5 & --- & 62.0 & --- \\
& Self-Critique & 82.0 & --- & 69.5 & --- & 63.7 & --- \\
& CoT & 75.5 & --- & 68.5 & --- & 78.8 & --- \\
\bottomrule
\end{tabular}
\end{table*}

For the archived Llama-3-8B runs, TRACE-LD point estimates are negative across the three benchmarks, and the HaluEval comparison of 500 paired questions has an interval excluding zero ($-5.8$pp, 95\% CI [$-$9.2, $-$2.4]). CoT gains +5.6pp on TriviaQA (500 questions, stored as aggregate accuracy only) and +12.0pp on 200 TruthfulQA questions. Prompting and Qwen results are otherwise reported as historical aggregates where per-answer verification is unavailable; in particular, a 200-question prompting score is not subtracted from the 500-question HaluEval greedy score. The logged Qwen2.5-7B scores point the same way for prompting on HaluEval QA and TriviaQA (CoT 68.5\% and 78.8\% against greedy 39.0\% and 62.8\%), but not on TruthfulQA (75.5\% against 76.1\%), and their sample counts were not recorded. The logs also show the TriviaQA CoT gain shrinking as greedy accuracy rises: about +16pp for Qwen2.5-7B (62.8\%), +5.6pp for Llama-3-8B (75.6\%) and +2pp for Llama-3.3-70B (90\%). These are three aggregates from one log, not an established trend.
These observations motivate matched cross-benchmark experiments rather than establish a universal ordering of methods.

\subsection{Mechanistic Analysis of Token-Level Interventions}
\label{sec:casestudy}

To understand \emph{why} token-level interventions have limited effect on instruction-tuned models, we select TRACE-LD as a representative case study. We choose it not because it is the strongest method (it is not), but because it is a typical learned logit adapter whose internal behavior can be inspected, providing diagnostics for this adapter rather than a mechanism established for every method in Section~\ref{sec:main}.

\paragraph{Token-level analysis.} Across 2{,}946 generated tokens, the adapter changes selections at 15.3\% of positions: 53\% function words, 45\% content words, uniformly distributed across response positions (34/32/34\% early/middle/late). The risk gate averages $\alpha\!=\!0.997$, nearly always fully active. This indicates diffuse, untargeted perturbation rather than focused factual correction.

\begin{figure}[t]
    \centering
    \includegraphics{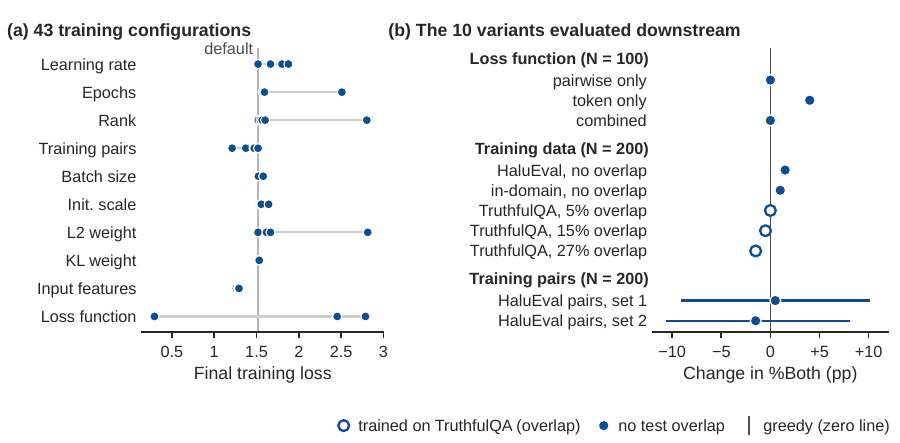}
    \caption{\textbf{Training losses and the variants evaluated downstream.} (a)~Final training loss of 43 TRACE-LD configurations; the light line marks the default; the three loss functions are different objectives, so their losses are not on a common scale. (b)~Change in \%Both against greedy (the dark zero line) for the ten variants evaluated downstream; as in Fig.~\ref{fig:sensitivities}, open circles mark training that overlaps the TruthfulQA test questions (without C1); whiskers: paired 95\% intervals.}
    \label{fig:ablation}
\end{figure}

\paragraph{Training ablation.} 43 configurations across 10 dimensions (learning rate, epochs, rank, data size, batch size, initialization, regularization, KL penalty, loss function, features) move the final training loss between 1.21 and 2.82 (Fig.~\ref{fig:ablation}a). We evaluated ten adapter variants downstream---three loss functions, five training sets, and two sets of training pairs (Fig.~\ref{fig:ablation}b). All of them land between $-1.5$ and $+4.0$pp of greedy, inside the sampling error of an $N\!=\!100$--$200$ evaluation, so we see no sign that further tuning of this adapter would change the picture.

\begin{figure}[t]
    \begin{minipage}[t]{0.485\textwidth}
        \centering
        \includegraphics{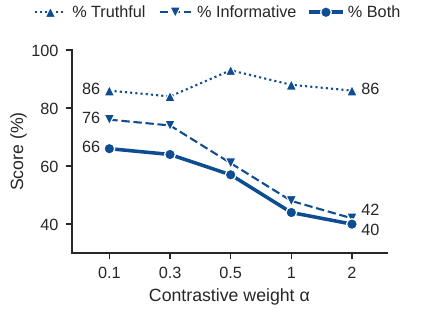}\\[-2pt]
        {\small (a)}
    \end{minipage}\hfill
    \begin{minipage}[t]{0.485\textwidth}
        \centering
        \includegraphics{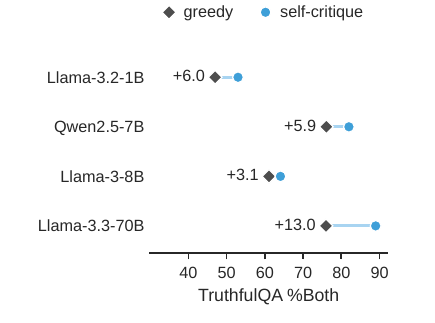}\\[-2pt]
        {\small (b)}
    \end{minipage}
    \caption{\textbf{Historical contrastive and self-critique observations.} (a)~TruthfulQA scores of Llama-3-8B-Instruct against the contrastive weight $\alpha$ (50 questions per setting, 100 at $\alpha\!=\!0.5$); the three labels of this token-level method share the blue of TRACE-LD and differ in line and marker. (b)~Greedy decoding (grey diamonds) against a second, self-critique pass (sky-blue circles, its colour in Fig.~\ref{fig:overview}) on four models; labels give the gain. Llama-3-8B: $N\!=\!817$; Llama-3.3-70B: $N\!=\!200$; the sample sizes of the other two runs were not recorded.}
    \label{fig:two}
\end{figure}

\paragraph{Contrastive decoding tradeoff.} Increasing the contrastive weight $\alpha$ from 0.1 to 2.0 lowers informativeness from 76\% to 42\%, while truthfulness stays between 84\% and 93\% without a trend (Fig.~\ref{fig:two}a). The net \%Both decreases from 66\% to 40\%: the method induces caution rather than improved factual accuracy.

\paragraph{DoLA layer sweep.} All 7 DoLA configurations (early/middle/late dynamic, every-4th, static-8/16/24) score below greedy (58.0--60.0\% vs.\ 61.0\%, $N\!=\!200$; Table~\ref{tab:app_dola}), indicating that layer-contrast decoding's +12pp on base models does not transfer to the instruction-tuned setting.

\subsection{Deliberative Methods}
\label{sec:deliberative}

In the original evaluated regime, deliberative prompting gave the largest gains of all tested methods. CoT and the revise-style variants gain 12.0--14.0pp on 200 TruthfulQA questions (Table~\ref{tab:delib}, Appendix~\ref{app:full}), against at most +2.0pp for the token-level methods of Table~\ref{tab:main}, and self-critique gains +3.1pp on all 817 questions ($p\!=\!.022$, Table~\ref{tab:full}). A fact-checker CoT prompt reaches 80.2\% on all 817 questions (+19.1pp, 95\% CI [+15.7, +22.5]); because it asks the model to check common beliefs, which is what TruthfulQA probes, we keep it out of the comparison tables (Appendix~\ref{app:cot}); Fig.~\ref{fig:overview} shows it, labelled as such, because it is the only CoT prompt with paired results on all 817 questions. We first describe these findings, then extend the framework to ask whether the self-critique gain transfers to additional judges on the same answers.

\begin{table}[t]
\centering
\caption{Deliberative method variants ($N\!=\!200$, TruthfulQA, Llama-8B). Single-pass CoT matches 2-pass self-critique, and rewriting without critique matches full critique-then-revise; these point estimates do not isolate reasoning as the causal mechanism.}
\label{tab:delib}
\small
\begin{tabular}{@{}lccc@{}}
\toprule
\textbf{Variant} & \textbf{Passes} & \textbf{\%Both} & \textbf{$\Delta$} \\
\midrule
Greedy & 1 & 61.0 & --- \\
\midrule
CoT prompting & 1 & 73.0 & +12.0 \\
Critique $\to$ revise & 2 & 73.5 & +12.5 \\
Rewrite w/o critique & 2 & 73.5 & +12.5 \\
Answer $\to$ explain & 2 & 59.5 & $-$1.5 \\
Verify only & 2 & 8.0 & $-$53.0 \\
\bottomrule
\end{tabular}
\end{table}

Three observations: (1) Single-pass CoT matches two-pass self-critique on Llama-3-8B (73.0\% vs.\ 73.5\%), showing that this score difference does not require two generation passes in that subset. (2) ``Rewrite without critique'' matches full ``critique then revise,'' showing that explicit critique is not isolated as the source of the observed gain. (3) ``Verify only'' fails catastrophically (8.0\%) because the model defaults to yes/no answers. The wording of the revision prompt matters as well: the self-critique prompt of the full benchmark (Table~\ref{tab:full}) scores 65.5\% on these 200 questions, 8pp below ``critique $\to$ revise.''

\paragraph{Does response length explain the CoT observation?} In the original evaluation, the concise-CoT variant scores 75.0\% on 200 questions with mean length 245 characters, compared with greedy's 61.0\% and 554 characters (Table~\ref{tab:app_cot}). This counters an explanation based solely on longer answers, but does not establish reasoning as the causal mechanism. The full-benchmark fact-checker CoT file retains scores only; its answer texts were not recovered, so it cannot be included in a fixed-answer cross-judge audit.

\paragraph{Across model sizes.} Self-critique gains reported in the original evaluation are +6.0pp (1B), +5.9pp (7B), +3.1pp (8B), and +13.0pp (70B; Fig.~\ref{fig:two}b). They are non-monotonic, use different question counts, and are not all backed by recoverable per-answer outputs. We treat these as exploratory observations rather than a scaling law. The exact judge provenance of the 70B run remains unverified; because Llama-3.3-70B also serves as the primary judge, self-judging or same-family preference is a concern, not an established property of that missing run.

\begin{table}[t]
\centering
\small
\caption{Self-critique versus its initial greedy answer on the same 817 questions. Intervals are paired 95\% bootstrap intervals; $p$ is an unadjusted exact McNemar test. The stored Llama marks come from the original run; for Qwen-235B and GPT-6 Astra, both arms were re-judged in the same new evaluation campaign.}
\label{tab:sc_rejudge}
\begin{tabular}{@{}lccccc@{}}
\toprule
Judge & Greedy & Revised & $\Delta$ & 95\% CI & $p$ \\
\midrule
Llama-3.3-70B (stored) & 61.1 & 64.1 & +3.1 & [+0.5, +5.6] & .022 \\
Qwen-235B & 50.9 & 53.6 & +2.7 & [+0.1, +5.3] & .053 \\
GPT-6 Astra & 39.4 & 35.5 & $-$3.9 & [$-$6.5, $-$1.5] & .004 \\
\bottomrule
\end{tabular}
\end{table}

\paragraph{Extension: transfer to additional judges.} The initial answer of every archived self-critique example is identical to the stored greedy answer. We re-score all 817 revised answers with Qwen-235B and GPT-6 Astra and obtain fresh greedy judgments in the same evaluation campaign (Table~\ref{tab:sc_rejudge}). The rubric is unchanged, with model-specific output budgets as in the TRACE-LD audit. Historical cached greedy judgments remain available as a separate repeatability comparison. Self-critique scores 53.6\% against Qwen's fresh greedy score of 50.9\% (+2.7pp, 95\% CI [+0.1, +5.3], exact $p\!=\!.053$). The percentile interval and discrete exact test give borderline different threshold decisions, so we report both. Under Astra, revisions score 35.5\% against 39.4\% for greedy: $-3.9$pp, CI [$-$6.5, $-$1.5], $p\!=\!.004$. The original +3.1pp remains the result under its original judge; its positive direction does not transfer to Astra in this added evaluation. With cached rather than fresh greedy marks, the differences are +3.1pp for Qwen and $-3.9$pp for Astra; the fresh and cached greedy \%Both labels agree on 97.9\% and 97.1\%, respectively. These are judgments of the same archived texts, not evidence that any judge supplies human ground truth. Section~\ref{sec:labels} examines which label and which revisions carry this difference.

\subsection{Where the Judges Disagree: Labels and Response Style}
\label{sec:labels}

\begin{figure}[t]
    \centering
    \includegraphics{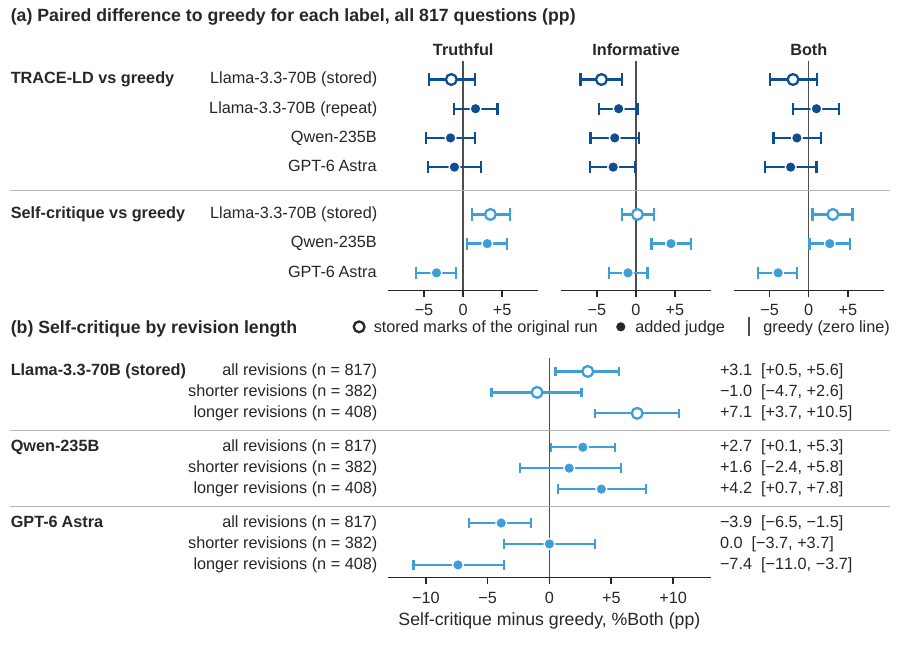}
    \caption{\textbf{Where the judges disagree.} Llama-3-8B-Instruct answers on all 817 TruthfulQA questions, each re-scored with the fixed answers; whiskers are paired 95\% bootstrap intervals. TRACE-LD in blue and self-critique in sky blue, as in Fig.~\ref{fig:overview}. Open circles: marks stored with the original runs; filled circles: added judges (the convention of the judge group in Fig.~\ref{fig:sensitivities}); the dark grey zero line is greedy. (a)~Differences to greedy for each of the rubric's two labels and for Both, under every judgment set. (b)~Self-critique minus greedy (\%Both) for all revisions and for revisions with fewer or more words than the initial answer (7 changed revisions of equal length and 20 unchanged ones are in neither subset). Exact values, and the other pre-specified subsets: Tables~\ref{tab:app_labels} and~\ref{tab:app_strata}.}
    \label{fig:labels}
\end{figure}

The transfer tests above summarize each answer by \%Both. Because the judges mark two labels and the answers are fixed, the stored records also show which label moves and in which answers. We fixed the following analyses from text properties before splitting any score by them, and report every subset (Appendix~\ref{app:edits}). They are descriptive: the subsets are not randomized, they overlap, and intervals are unadjusted.

\paragraph{Which label moves.} For TRACE-LD, the informativeness difference is negative under all four judgment sets ($-4.4$pp, 95\% CI [$-$7.1, $-$1.8], with the stored marks; $-2.2$ to $-2.9$ with the others, two of whose intervals include zero), whereas the truthfulness difference changes sign ($-1.6$ to $+1.6$; Fig.~\ref{fig:labels}a). The four sets score the same answers and are not independent replications; still, the direction they share is a small loss of informativeness, not a gain in truthfulness. For self-critique, the judges diverge on truthfulness: $+3.5$pp [+1.1, +6.0] with the stored Llama marks, $+3.1$ [+0.5, +5.6] under Qwen and $-3.4$ [$-$6.0, $-$0.9] under Astra. On the TRACE-LD answers, agreement with the stored marks is lower for informativeness than for truthfulness (Cohen's $\kappa$ 0.44 vs.\ 0.69 for Qwen and 0.29 vs.\ 0.38 for Astra), and the two added judges agree with each other at $\kappa\!=\!0.49$ on truthfulness and 0.47 on informativeness. Disagreement with the original judge is therefore not a consensus of the newer judges against it.

\paragraph{Which revisions.} The archived self-critique script generates the initial answer with at most 128 new tokens and the revision with at most 256, and asks for the revised answer ``without meta-commentary''. Of the 817 initial answers, 595 stop mid-sentence, against 99 of the revisions; 236 revisions contain meta-text such as ``Here is the revised answer:'', and 307 are longer than the longest initial answer (118 words). On the 382 revisions with fewer words than their initial answer, no judge detects a difference ($-1.0$, $+1.6$ and $0.0$pp). On the 408 longer revisions, the stored Llama marks give $+7.1$pp [+3.7, +10.5], Qwen $+4.2$ [+0.7, +7.8] and Astra $-7.4$ [$-$11.0, $-$3.7] (Fig.~\ref{fig:labels}b). The Llama effect exceeds the Astra effect by $+14.5$pp [+9.8, +19.4] on longer revisions and by $-1.0$pp [$-$5.5, +3.4] on shorter ones. Revisions with meta-text and revisions that complete a cut-off answer show the same split (Table~\ref{tab:app_strata}). These features largely coincide---220 of the 236 revisions with meta-text are longer---so the data cannot separate length, completion, meta-text and generation budget.

\paragraph{What this does and does not show.} Without human labels, these results do not show which judge is right about the longer revisions, and the subsets do not establish a causal effect of length. They do locate the transfer boundary of Sect.~\ref{sec:deliberative}: the original self-critique gain and its reversal under Astra both arise mainly in the longer revisions---307 of the 408 exceed the length that any initial answer reached under its 128-token limit---and mostly in the truthfulness label. One hypothesis consistent with this pattern is that a rubric under which any false claim makes an answer untruthful gives longer answers more chances to fail under a strict judge; testing it requires claim-level or human annotation. For the framework, two consequences follow: generation budgets belong among the matched conditions of C3 and C4, and a judge-transfer test should be read per label and per response style rather than only through \%Both. Seventeen revisions are byte-identical to their initial answers, so each added judge received these 17 inputs twice within one campaign; each judge changed one label on one of them (Appendix~\ref{app:edits}). Temperature-0 scoring through the gateway is therefore not exactly repeatable.

\subsection{Evaluation Checklist}
\label{sec:checklist}

The seven-point checklist (Table~\ref{tab:checklist}) converts the five sensitivities into reporting requirements. This extension connects those requirements to stored answer identifiers, model/run metadata, paired statistics, and source-linked artifacts, and adds two requirements supported by Sect.~\ref{sec:labels}: per-label reporting of judge effects (item~2) and matched generation budgets (item~4). Its purpose is to make both positive results and limits on their transfer independently inspectable.

\begin{table}[t]
\centering
\caption{Evaluation checklist for truthfulness research. Each item is phrased as an executable requirement tied to a specific sensitivity.}
\label{tab:checklist}
\small
\begin{tabularx}{\textwidth}{@{}cXc@{}}
\toprule
\textbf{\#} & \textbf{Requirement} & \textbf{Ref.} \\
\midrule
1 & Report train/eval overlap, dataset versions and the relevant example identifiers. & S1 \\
2 & Use multiple judge families; report per-judge effects for each rubric label, agreement and model/run identifiers. & S2 \\
3 & Test $\geq$2 judge prompt variants on fixed outputs; retain the prompts and answer identifiers. & S2 \\
4 & Compare against $T$/top-$p$ sweeps under matched conditions, question sets and generation budgets (per pass). & S3 \\
5 & Report length/refusal/meta-text analyses separately, including exact filters and subset sizes. & S4 \\
6 & Provide paired CIs ($\geq$1{,}000 bootstrap resamples), seeds and declared testing procedures. & S5 \\
7 & Report repeated evaluations ($\geq$3 seeds where applicable), the varying factors and failure handling. & S5 \\
\bottomrule
\end{tabularx}
\end{table}

\section{Discussion and Conclusion}
\label{sec:discussion}

\paragraph{Evaluation methodology as a first-class concern.}
The six-control framework is the common contribution across the original study and this extension. The original experiments identify settings in which apparent gains depend on evaluation choices; the added fixed-answer tests make judge dependence and repeatability directly inspectable. TRACE-LD has no detected positive gain across the collected judgment sets, while absolute scores shift substantially. The label-level and response-style analyses add a further reason to keep the controls together rather than apply them one at a time: judge choice (C2) and response style (C4) interact, so a judge-transfer test summarized only by \%Both can hide where, and in which label, two evaluators part ways. The framework therefore supports reporting both the original-regime result and the extent to which that result transfers to another evaluator.

\paragraph{Token-level methods vs.\ deliberative approaches.}
In the original evaluated regime, token-level methods show limited gains, none of them significant, while deliberative prompting gives the largest: on Llama-3-8B-Instruct, CoT gains +12.0pp on 200 TruthfulQA questions and +5.6pp on TriviaQA, a fact-checker CoT prompt +19.1pp on all 817 TruthfulQA questions, and self-critique +3.1pp. One interpretation, which these experiments do not test, is that instruction tuning already captures much of what output-distribution adjustments can reach, so that the remaining errors respond more to how the model reasons than to logit perturbations. These are statements about the tested models, prompts, benchmarks and judges, not guarantees under every evaluator. The added self-critique experiment locates a specific transfer boundary: Astra assigns a negative paired effect to revisions that received a positive effect from the original judge. The boundary lies mostly in the truthfulness label and in revisions longer than their initial answer, which the archived protocol generated with twice the token budget; on shorter revisions none of the three judges detects a difference. A comparison of self-critique with a budget-matched baseline is therefore a natural next experiment. The original observation and the extension answer different questions and are reported separately. This strengthens the motivation for judge-transfer checks without establishing a universal method ordering or a uniquely correct judge. CoT has not received the same fixed-answer re-judging because its answer texts were not recovered; its results stand under the original judge and are neither confirmed nor contradicted by the transfer tests.

\paragraph{Practical guidance.}
The practical output is the seven-point checklist together with an executable evidence path: stored answers, judge records, derived statistics and checked figures/tables. Researchers can use it to retain an original result with its protocol while documenting transfer tests, uncertainty and failures. Because judges mark several labels, storing every label rather than only an aggregate pass costs nothing at evaluation time and is what made the analysis of Sect.~\ref{sec:labels} possible after the fact. Comparisons of training overlap or prompting mechanisms still require matched experiments; those future tests extend the framework rather than replace its existing contributions.

\paragraph{Scope boundaries and threats to validity.}
Our conclusions are conditional on the tested models, question sets, generated answers and scoring rubrics. We have no human labels, so disagreement cannot identify which judge is correct. The original generation scores are Llama-3.3-70B judgments except where another judge is explicitly named. The re-judged greedy and TRACE-LD scores are 10--22 points lower under Qwen and Astra than under the stored Llama marks. API model identifiers may refer to changing deployments, and the historical evaluator could fall back to a heuristic after errors, unlike the new scoring script. The new self-critique audit collects both arms in the same evaluation campaign, with revisions and baselines scored in successive batches rather than interleaved. Earlier cached greedy judgments are retained as a repeatability comparison. Method-comparison $p$ values are nominal and unadjusted across the exploratory inventory; lack of significance does not establish equivalence. The label-level and response-style subsets are descriptive and overlapping; they were fixed from text properties before scores were split, but they are not randomized comparisons. DoLA/ITI are reimplementations, and several small-subset or log-only experiments cannot support broad causal or scaling claims.

\paragraph{Conclusion.}
We contribute a six-control framework, five empirical evaluation sensitivities, a study of 15 methods on five models (1B--70B) and three benchmarks, and a seven-point checklist. On these instruction-tuned models, token-level decoding methods show limited gains in the original evaluated regime: none of the learned or prior methods in the main comparison exceeds a simple temperature baseline, and the best learned adapter stays below greedy on TruthfulQA and HaluEval QA. Deliberative prompting, which needs no training, gives the largest gains. This version retains these findings and extends the evidence with archived full-benchmark judge-transfer experiments, label-level and response-style analyses, and reproducibility tools. The extended results distinguish two outcomes: TRACE-LD remains without a detected positive gain under every evaluated judge, with a small informativeness loss as the direction shared by all judgment sets, while the original self-critique gain does not transfer to Astra; that disagreement lies mainly in the truthfulness label and in longer revisions produced under a larger generation budget. CoT could not be re-judged. Together, the original study and the extension show how to report method effectiveness with its measurement conditions and how to test the reach of a result beyond those conditions.

\bibliographystyle{splncs04}
\phantomsection\addcontentsline{toc}{section}{References} 
\bibliography{references}

\begin{thebibliography}{10}
\providecommand{\url}[1]{\texttt{#1}}
\providecommand{\urlprefix}{URL }
\providecommand{\doi}[1]{https://doi.org/#1}

\bibitem{bai_training_2022}
Bai, Y., et~al.: Training a {Helpful} and {Harmless} {Assistant} with {Reinforcement} {Learning} from {Human} {Feedback} (Apr 2022)

\bibitem{chiang_can_2023}
Chiang, C.H., Lee, H.y.: Can {Large} {Language} {Models} {Be} an {Alternative} to {Human} {Evaluations}? (May 2023)

\bibitem{christiano_deep_2023}
Christiano, P., Leike, J., Brown, T.B., Martic, M., Legg, S., Amodei, D.: Deep reinforcement learning from human preferences (Feb 2023)

\bibitem{chuang_dola_2024}
Chuang, Y.S., Xie, Y., Luo, H., Kim, Y., Glass, J., He, P.: {DoLa}: {Decoding} by {Contrasting} {Layers} {Improves} {Factuality} in {Large} {Language} {Models} (Mar 2024)

\bibitem{chung_scaling_2022}
Chung, H.W., et~al.: Scaling {Instruction}-{Finetuned} {Language} {Models} (Dec 2022)

\bibitem{hernandez_inspecting_2024}
Hernandez, E., Li, B.Z., Andreas, J.: Inspecting and {Editing} {Knowledge} {Representations} in {Language} {Models} (Aug 2024)

\bibitem{li_halueval_2023}
Li, J., Cheng, X., Zhao, W.X., Nie, J.Y., Wen, J.R.: {HaluEval}: {A} {Large}-{Scale} {Hallucination} {Evaluation} {Benchmark} for {Large} {Language} {Models} (Oct 2023)

\bibitem{li_inference-time_2024}
Li, K., Patel, O., Viégas, F., Pfister, H., Wattenberg, M.: Inference-{Time} {Intervention}: {Eliciting} {Truthful} {Answers} from a {Language} {Model} (Jun 2024)

\bibitem{li_contrastive_2023}
Li, X.L., Holtzman, A., Fried, D., Liang, P., Eisner, J., Hashimoto, T., Zettlemoyer, L., Lewis, M.: Contrastive {Decoding}: {Open}-ended {Text} {Generation} as {Optimization} (Jul 2023)

\bibitem{lin_truthfulqa_2022}
Lin, S., Hilton, J., Evans, O.: {TruthfulQA}: {Measuring} {How} {Models} {Mimic} {Human} {Falsehoods} (May 2022)

\bibitem{obrien_contrastive_2023}
O'Brien, S., Lewis, M.: Contrastive {Decoding} {Improves} {Reasoning} in {Large} {Language} {Models} (Sep 2023)

\bibitem{ouyang_training_2022}
Ouyang, L., et~al.: Training language models to follow instructions with human feedback (Mar 2022)

\bibitem{panickssery_llm_2024}
Panickssery, A., Bowman, S.R., Feng, S.: {LLM} {Evaluators} {Recognize} and {Favor} {Their} {Own} {Generations} (Apr 2024)

\bibitem{qwen_qwen25_2025}
Qwen, Yang, A., Yang, B., et~al.: Qwen2.5 {Technical} {Report} (Jan 2025)

\bibitem{schaeffer_are_2023}
Schaeffer, R., Miranda, B., Koyejo, S.: Are {Emergent} {Abilities} of {Large} {Language} {Models} a {Mirage}? (May 2023)

\bibitem{sun_chair_2025}
Sun, A.: {CHAIR} -- {Classifier} of {Hallucination} as {Improver} (Jan 2025)

\bibitem{turner_steering_2024}
Turner, A.M., Thiergart, L., Leech, G., Udell, D., Vazquez, J.J., Mini, U., MacDiarmid, M.: Steering {Language} {Models} {With} {Activation} {Engineering} (Oct 2024)

\bibitem{wang_large_2023}
Wang, P., et~al.: Large {Language} {Models} are not {Fair} {Evaluators} (Aug 2023)

\bibitem{wu_style_2023}
Wu, M., Aji, A.F.: Style {Over} {Substance}: {Evaluation} {Biases} for {Large} {Language} {Models} (Nov 2023)

\bibitem{zheng_judging_2023}
Zheng, L., et~al.: Judging {LLM}-as-a-{Judge} with {MT}-{Bench} and {Chatbot} {Arena} (Dec 2023)

\end{thebibliography}

\appendix

\section{Complete Method Comparison}
\label{app:full}

Table~\ref{tab:app_full} provides the complete comparison of the methods evaluated on Llama-3-8B-Instruct, including compute cost estimates. Sample sizes differ between experiments, so every $\Delta$ is computed against greedy decoding on the same questions in the same run.

\begin{table}[H]
\centering
\caption{Complete method comparison on Llama-3-8B-Instruct (TruthfulQA). $N$: number of questions; Greedy: \%Both of greedy decoding on those questions.}
\label{tab:app_full}
\small
\begin{tabular}{@{}llccccc@{}}
\toprule
\textbf{Category} & \textbf{Method} & $N$ & \textbf{Greedy} & \textbf{\%Both} & \textbf{$\Delta$} & \textbf{Compute} \\
\midrule
\multirow{2}{*}{Simple} & $T\!=\!0.3$ & 817 & 61.1 & 62.1 & +1.0 & $1\times$ \\
& top-$p\!=\!0.5$ & 100 & 59.0 & 61.0 & +2.0 & $1\times$ \\
\midrule
\multirow{3}{*}{Prior} & DoLA (best of 7) & 200 & 61.0 & 60.0 & $-$1.0 & $\sim$1.3$\times$ \\
& ITI (best of 4) & 100 & 58.0 & 56.0 & $-$2.0 & $1\times$ \\
& Contrastive ($\alpha\!=\!0.5$) & 100 & 59.0 & 57.0 & $-$2.0 & $1\times$ \\
\midrule
\multirow{3}{*}{Learned} & TRACE-LD (2.3M) & 817 & 61.1 & 59.1 & $-$2.0 & $1\times$ \\
& HiddenSteer & 200 & 56.5 & 52.5 & $-$4.0 & $1\times$ \\
& LoRA (6.8M) & 100 & 59.0 & 49.0 & $-$10.0 & $1\times$ \\
\midrule
\multirow{5}{*}{Deliberative} & CoT prompting & 200 & 61.0 & 73.0 & +12.0 & $\sim$1.5$\times$ \\
& Concise CoT & 200 & 61.0 & 75.0 & +14.0 & $\sim$0.5$\times$ \\
& Critique $\to$ revise & 200 & 61.0 & 73.5 & +12.5 & $2\times$ \\
& Self-critique & 817 & 61.1 & 64.1 & +3.1 & $2\times$ \\
& $T\!=\!0.3$ + CoT & 200 & 61.0 & 75.0 & +14.0 & $\sim$1.5$\times$ \\
\bottomrule
\end{tabular}
\end{table}

\section{DoLA Layer Sweep}
\label{app:dola}

We sweep 7 DoLA layer configurations on Llama-3-8B-Instruct ($N\!=\!200$). All configurations score below greedy. Our implementation follows the published algorithm; the official codebase requires a patched \texttt{transformers} 4.28.1 that does not support Llama-3's grouped-query attention.

\begin{table}[H]
\centering
\caption{DoLA layer sweep. No configuration improves over greedy ($\sim$61\%).}
\label{tab:app_dola}
\small
\begin{tabular}{@{}lccc@{}}
\toprule
\textbf{Config} & \textbf{Premature Layers} & \textbf{\%Both} & \textbf{$\Delta$} \\
\midrule
Early dynamic & 0--15 & 58.0 & $-$3.0 \\
Middle dynamic & 8--23 & 59.5 & $-$1.5 \\
Late dynamic & 16--31 & 60.0 & $-$1.0 \\
Every-4th & 0, 4, ..., 28 & 58.5 & $-$2.5 \\
Static (layer 8) & 8 vs 32 & 60.0 & $-$1.0 \\
Static (layer 16) & 16 vs 32 & 60.0 & $-$1.0 \\
Static (layer 24) & 24 vs 32 & 58.0 & $-$3.0 \\
\bottomrule
\end{tabular}
\end{table}

\section{Judge Models and Judge Prompts}
\label{app:judge}

Table~\ref{tab:app_judges} transcribes a historical 100-question judge comparison. Its complete per-answer artifacts have not been recovered; it is not the fixed-answer comparison used in Fig.~\ref{fig:sensitivities}. Table~\ref{tab:app_prompts} shows the effect of judge prompt wording on identical greedy outputs ($N\!=\!100$). Scores range from 76\% to 83\% depending only on prompt phrasing.

\begin{table}[H]
\centering
\caption{Historical log-only judge comparison ($N\!=\!100$, \%Both). Input identity cannot currently be independently checked from recovered per-answer artifacts.}
\label{tab:app_judges}
\small
\begin{tabular}{@{}lccc@{}}
\toprule
\textbf{Judge Configuration} & \textbf{Greedy} & \textbf{TRACE-LD} & \textbf{$\Delta$} \\
\midrule
Llama-3.3-70B (default prompt) & 59.0 & 60.0 & +1.0 \\
Llama-3.3-70B (rubric prompt) & 83.0 & 80.0 & $-$3.0 \\
Qwen-235B & 70.0 & 49.0 & $-$21.0 \\
\bottomrule
\end{tabular}
\end{table}

Table~\ref{tab:app_rejudge} repeats the comparison at full scale on answers that are stored: the 817 greedy and 817 TRACE-LD answers behind Table~\ref{tab:full}, scored independently by three judge models using the archived rubric and temperature 0. Qwen and Llama use a 64-token limit; Astra uses 2048 because 64 tokens sometimes exhausted its reasoning budget before a score was emitted. New runs retry failures instead of using the historical evaluator's heuristic fallback. Stored and repeated Llama marks are two sets from the same model identifier, not independent judge families.

\begin{table}[H]
\centering
\caption{The full-benchmark comparison under four sets of judge marks ($N\!=\!817$, \%Both). Stored: the marks kept with the runs, which the paper uses. Agree, $\kappa$: share of the 1634 answers marked as in the stored marks, and Cohen's $\kappa$.}
\label{tab:app_rejudge}
\small
\begin{tabular}{@{}lccccccc@{}}
\toprule
\textbf{Judge} & \textbf{Greedy} & \textbf{TRACE-LD} & \textbf{$\Delta$} & \textbf{95\% CI} & \textbf{$p$} & \textbf{Agree} & \textbf{$\kappa$} \\
\midrule
Llama-3.3-70B (stored) & 61.1 & 59.1 & $-$2.0 & [$-$4.9, +1.1] & .24 & --- & --- \\
Llama-3.3-70B (run again) & 63.2 & 64.1 & +1.0 & [$-$2.0, +3.9] & .57 & 94.5 & 0.88 \\
Qwen-235B & 50.6 & 49.1 & $-$1.5 & [$-$4.5, +1.6] & .39 & 85.2 & 0.70 \\
GPT-6 Astra & 39.4 & 37.1 & $-$2.3 & [$-$5.6, +1.0] & .20 & 67.6 & 0.38 \\
\bottomrule
\end{tabular}
\end{table}

\begin{table}[H]
\centering
\caption{Judge prompt sensitivity on identical outputs.}
\label{tab:app_prompts}
\small
\begin{tabular}{@{}lcccc@{}}
\toprule
\textbf{Prompt Variant} & \textbf{\%Truth} & \textbf{\%Info} & \textbf{\%Both} & \textbf{$\Delta$} \\
\midrule
Rubric (detailed scoring) & 89.0 & 84.0 & 83.0 & --- \\
Binary (yes/no) & 86.0 & 87.0 & 80.0 & $-$3.0 \\
Strict (harsh criteria) & 79.0 & 85.0 & 76.0 & $-$7.0 \\
\bottomrule
\end{tabular}
\end{table}

\section{Label-Level and Response-Style Analyses}
\label{app:edits}

This appendix gives the rules and complete results behind Sect.~\ref{sec:labels}. All inputs are archived: the stored marks of the full-benchmark runs, the new judge records of both campaigns, and the self-critique script in the code snapshot, from which the generation budgets (128 new tokens for the initial answer, 256 for the revision) are read programmatically. The rules were written down from text properties alone before any score was split by them; every subset defined in advance is reported in Table~\ref{tab:app_strata}.

\paragraph{Rules.} Words are whitespace-separated tokens. A revision is \emph{unchanged} if it equals its initial answer after collapsing whitespace (20 revisions; 17 are byte-identical), \emph{shorter} or \emph{longer} if it is changed and has fewer or more words (382 and 408; 7 changed revisions keep the word count). It has \emph{meta-text} if it matches the case-insensitive pattern \texttt{revised|\textbackslash bnote:} (236). An answer is \emph{cut off} if it does not end in one of \texttt{. ! ? " ' ) *}; a revision \emph{completes a cut-off answer} if its initial answer is cut off and it is not (525).

\paragraph{Labels.} Table~\ref{tab:app_labels} lists the paired differences for each label. Agreement is computed over the pooled greedy and method answers of a comparison (1,634 answers). On the TRACE-LD answers, agreement with the stored marks is 85.0\% (truthful) and 81.4\% (informative) for Qwen-235B and 67.1\% and 70.3\% for GPT-6 Astra; the two added judges agree with each other on 73.9\% and 76.6\%. The self-critique answers give similar values.

\begin{table}[H]
\centering
\footnotesize
\caption{Label-level paired differences to greedy (pp) on the fixed answers of all 817 TruthfulQA questions, with paired 95\% bootstrap intervals. Truthful and Informative are the two marks of the rubric; Both requires both.}
\label{tab:app_labels}
\setlength{\tabcolsep}{3.5pt}
\begin{tabular}{@{}lccc@{}}
\toprule
Judge & Truthful & Informative & Both \\
\midrule
\multicolumn{4}{@{}l}{\textit{TRACE-LD vs.\ greedy}} \\
Llama-3.3-70B (stored) & $-$1.5 [$-$4.4, +1.5] & $-$4.4 [$-$7.1, $-$1.8] & $-$2.0 [$-$4.9, +1.1] \\
Llama-3.3-70B (repeat) & +1.6 [$-$1.2, +4.4] & $-$2.2 [$-$4.7, +0.2] & +1.0 [$-$2.0, +3.9] \\
Qwen-235B & $-$1.6 [$-$4.7, +1.5] & $-$2.7 [$-$5.8, +0.4] & $-$1.5 [$-$4.5, +1.6] \\
GPT-6 Astra & $-$1.1 [$-$4.5, +2.3] & $-$2.9 [$-$5.9, $-$0.1] & $-$2.3 [$-$5.6, +1.0] \\
\midrule
\multicolumn{4}{@{}l}{\textit{Self-critique vs.\ greedy}} \\
Llama-3.3-70B (stored) & +3.5 [+1.1, +6.0] & +0.2 [$-$1.8, +2.3] & +3.1 [+0.5, +5.6] \\
Qwen-235B & +3.1 [+0.5, +5.6] & +4.5 [+2.0, +7.1] & +2.7 [+0.1, +5.3] \\
GPT-6 Astra & $-$3.4 [$-$6.0, $-$0.9] & $-$1.0 [$-$3.4, +1.5] & $-$3.9 [$-$6.5, $-$1.5] \\
\bottomrule
\end{tabular}
\end{table}

\begin{table}[H]
\centering
\footnotesize
\caption{Self-critique minus greedy (\%Both, pp) within subsets of the 817 revisions defined by text rules alone (Appendix~\ref{app:edits}); paired 95\% bootstrap intervals. Subsets overlap; all pre-specified subsets are shown.}
\label{tab:app_strata}
\setlength{\tabcolsep}{3.5pt}
\begin{tabular}{@{}>{\raggedright\arraybackslash}p{2.45cm}rccc@{}}
\toprule
Revision subset & $n$ & Llama (stored) & Qwen-235B & GPT-6 Astra \\
\midrule
all revisions & 817 & +3.1 [+0.5, +5.6] & +2.7 [+0.1, +5.3] & $-$3.9 [$-$6.5, $-$1.5] \\
unchanged & 20 & 0.0 [0.0, 0.0] & 0.0 [0.0, 0.0] & $-$5.0 [$-$15.0, 0.0] \\
shorter & 382 & $-$1.0 [$-$4.7, +2.6] & +1.6 [$-$2.4, +5.8] & 0.0 [$-$3.7, +3.7] \\
longer & 408 & +7.1 [+3.7, +10.5] & +4.2 [+0.7, +7.8] & $-$7.4 [$-$11.0, $-$3.7] \\
with meta-text & 236 & +9.7 [+5.1, +14.4] & +5.5 [+0.8, +10.2] & $-$8.1 [$-$12.7, $-$3.4] \\
changed, no meta-text & 561 & +0.4 [$-$2.7, +3.4] & +1.6 [$-$1.6, +4.8] & $-$2.1 [$-$5.3, +1.1] \\
completes a cut-off answer & 525 & +4.4 [+1.1, +7.6] & +3.0 [$-$0.2, +6.3] & $-$5.3 [$-$8.8, $-$1.9] \\
changed, does not complete one & 272 & +0.7 [$-$3.3, +4.8] & +2.2 [$-$2.6, +7.0] & $-$1.1 [$-$5.5, +3.3] \\
\bottomrule
\end{tabular}
\end{table}

\paragraph{Identical inputs.} For the 17 byte-identical revisions, the new judges received the same request twice within one campaign (identical input hashes). Qwen-235B changed the informativeness label of one of these answers and GPT-6 Astra the truthfulness label of another; the stored Llama marks of the original run agree on all 17. Seventeen pairs cannot estimate a repeat-disagreement rate; they only show that the rate is not zero, in line with the 97.9\% and 97.1\% agreement of fresh and cached greedy marks (Sect.~\ref{sec:deliberative}).

\section{Seed Variance, Confounds, and Contamination}
\label{app:seeds}

Table~\ref{tab:app_seeds} shows greedy baseline variance across 3 evaluation seeds, describing variability in those three runs rather than a universal noise floor. Table~\ref{tab:app_confounds} shows the effect of refusal and length controls on TRACE-LD's apparent gain, and Table~\ref{tab:app_contam} the contamination dose-response.

\begin{table}[H]
\centering
\caption{Seed variance on greedy baseline ($N\!=\!100$). The last row is transcribed from the experiment log; computed from the three values shown, the \%Both standard deviation is 1.2pp (population) or 1.5pp (sample).}
\label{tab:app_seeds}
\small
\begin{tabular}{@{}lccc@{}}
\toprule
\textbf{Seed} & \textbf{\%Truth} & \textbf{\%Info} & \textbf{\%Both} \\
\midrule
42 & 66.0 & 70.0 & 59.0 \\
123 & 69.0 & 76.0 & 61.0 \\
456 & 67.0 & 69.0 & 58.0 \\
\midrule
Mean $\pm$ Std & 67.3 $\pm$ 1.3 & 71.7 $\pm$ 3.1 & 59.3 $\pm$ 1.3 \\
\bottomrule
\end{tabular}
\end{table}

\begin{table}[H]
\centering
\caption{Confound controls on TRACE-LD (first 200 TruthfulQA questions, recomputed from per-question judgments; 95\% paired-bootstrap intervals). The adapter's answers average 55.3 tokens against 30.8 for greedy, and 1.0\% of them contain the refusal ``I have no comment'' against 7.5\%.}
\label{tab:app_confounds}
\small
\begin{tabular}{@{}lccccc@{}}
\toprule
\textbf{Questions} & $n$ & \textbf{Greedy} & \textbf{TRACE-LD} & \textbf{$\Delta$} & \textbf{95\% CI} \\
\midrule
All & 200 & 59.0 & 59.5 & +0.5 & [$-$9.0, +10.0] \\
Greedy does not refuse & 185 & 63.8 & 61.1 & $-$2.7 & [$-$11.9, +7.0] \\
Adapter answer $\leq$ 40 tokens & 75 & 57.3 & 58.7 & +1.3 & [$-$14.7, +17.3] \\
\bottomrule
\end{tabular}
\end{table}

\begin{table}[H]
\centering
\caption{Contamination dose-response (logits-only adapter, $N\!=\!200$). These reported point estimates do not isolate a contamination mechanism for the full adapter.}
\label{tab:app_contam}
\small
\begin{tabular}{@{}lccc@{}}
\toprule
\textbf{Training Data} & \textbf{Overlap} & \textbf{\%Both} & \textbf{$\Delta$} \\
\midrule
None (Greedy) & --- & 61.0 & --- \\
HaluEval (OOD, 512 ex.) & 0\% & 62.5 & +1.5 \\
In-domain, no overlap & 0\% & 62.0 & +1.0 \\
TruthfulQA, 5\% overlap & 5\% & 61.0 & $\pm$0 \\
TruthfulQA, 15\% overlap & 15\% & 60.5 & $-$0.5 \\
TruthfulQA, 27\% overlap & 27\% & 59.5 & $-$1.5 \\
\bottomrule
\end{tabular}
\end{table}

\section{CoT Prompt Robustness}
\label{app:cot}

Table~\ref{tab:app_cot} shows that CoT improvements are robust across different prompt wordings, and that ``concise CoT'' achieves comparable gains with shorter responses than greedy. The three wordings were compared on the first 100 questions, the length control on the first 200; each $\Delta$ is against the standard prompt on the same questions.

\begin{table}[H]
\centering
\caption{CoT prompt variants and length control.}
\label{tab:app_cot}
\small
\begin{tabular}{@{}lccc@{}}
\toprule
\textbf{Prompt Variant} & \textbf{\%Both} & \textbf{Avg Len (chars)} & \textbf{$\Delta$} \\
\midrule
\multicolumn{4}{l}{\textit{First 100 questions}} \\
Standard prompt & 69.0 & --- & --- \\
CoT: ``step by step'' & 83.0 & 1144 & +14.0 \\
CoT: ``consider misconceptions'' & 82.0 & --- & +13.0 \\
CoT: ``fact-checker persona'' & 93.0 & --- & +24.0 \\
\midrule
\multicolumn{4}{l}{\textit{First 200 questions}} \\
Greedy (standard) & 61.0 & 554 & --- \\
Concise CoT (max 64 tok) & 75.0 & 245 & +14.0 \\
\bottomrule
\end{tabular}
\end{table}

\noindent Note: The ``fact-checker persona'' prompt achieves 93\% on the first 100 questions and 80.2\% on the full benchmark ($N\!=\!817$; +19.1pp over greedy, 95\% CI [+15.7, +22.5]). It may benefit from prompt--benchmark alignment, as TruthfulQA specifically tests common misconceptions and this prompt explicitly instructs the model to evaluate common beliefs; we therefore do not report it among the main comparison results, and Fig.~\ref{fig:overview} labels it as the fact-checker prompt. The more conservative ``concise CoT'' result (+14.0pp at 245 chars vs.\ greedy's 554 chars) is inconsistent with a gain caused solely by longer answers, but does not establish a reasoning mechanism. The full-benchmark fact-checker file contains scores without the generated answer texts; those outputs were not recovered for independent re-judging.

\section{TRACE-LD Training Details}
\label{app:trace}

TRACE-LD is a low-rank logit correction adapter with the following architecture:
\begin{itemize}[leftmargin=*,itemsep=1pt]
    \item \textbf{Input features:} model hidden states (4096-dim) + top-50 logit statistics (55-dim)
    \item \textbf{Feature projection:} 4151 $\to$ 64 (two-layer MLP with GELU)
    \item \textbf{Low-rank factorization:} rank $r\!=\!16$; gate head (64 $\to$ 16), basis (16 $\to$ $V$)
    \item \textbf{Risk gate:} sigmoid-gated scaling $\alpha \in (0, 1)$
    \item \textbf{Delta clamping:} logit corrections clamped to $\pm 2.0$
    \item \textbf{Total parameters:} 2.3M trainable (model frozen)
\end{itemize}

\noindent Training: AdamW optimizer, learning rate $5 \times 10^{-4}$, batch size 2, 5 epochs on 512 HaluEval QA pairs. Training time: $\sim$10 minutes on a single A100 GPU.

\section{Reproducibility}
\label{app:repro}

The generation experiments use publicly available models (Llama-3-8B-Instruct, Qwen2.5-7B-Instruct from HuggingFace) and datasets (TruthfulQA, HaluEval, TriviaQA from HuggingFace Datasets). LLM judge calls use the NVIDIA Inference API (Llama-3.3-70B-Instruct, Qwen-235B and GPT-6 Astra). The project archive holds the experiment code snapshot (37 scripts), the per-question outputs and judge marks of the full-benchmark runs, 20 run folders with predictions and configurations, the training histories of 64 adapters, 24 experiment logs, and all 8,170 new judge records; the 64 adapter checkpoints and the per-token traces are kept separately because of their size. We will release the code, these outputs and the adapter checkpoints. Figures combine per-question outputs, training histories and explicitly identified log-only aggregates. The full-benchmark intervals, tests and generated tables are recomputed from archived marks by one script, which also reads the generation budgets from the archived experiment code; a second script checks the numbers in Fig.~\ref{fig:overview}, the generated tables and the sentences that quote them. New self-critique records additionally retain input and protocol hashes, output limits, raw scoring replies and retry metadata; gateway aliases do not establish immutable model versions.

\section{Coverage and Artifact Availability}
\label{app:coverage}

The inventory of five models, three benchmarks and 15 methods is not a complete factorial experiment: the 1B and 70B models, for example, were not evaluated with every method or benchmark. Artifact availability differs by comparison, as summarized below. Historical results remain in the appendix for traceability and should not be treated as independently reproduced findings.

\begin{table}[H]
\centering
\small
\caption{Evidence coverage. Full-benchmark re-judging keeps the generated answers fixed; other rows have narrower or incomplete coverage.}
\label{tab:app_coverage}
\begin{tabular}{@{}>{\raggedright\arraybackslash}p{3.2cm}>{\raggedright\arraybackslash}p{5.2cm}>{\raggedright\arraybackslash}p{5.9cm}@{}}
\toprule
Comparison & Available artifacts & Scope / limitation \\
\midrule
Greedy vs.\ TRACE-LD & 817 answers per arm; original marks; three new judge runs & Llama-3-8B-Instruct, TruthfulQA; no human labels \\
Self-critique & 817 initial and revised answers; original marks; Qwen/Astra re-judging & Both new-judge arms scored in one campaign; archived earlier baseline marks also retained \\
Label and response-style analyses & Both labels of every stored mark; revision texts; archived generation script & Descriptive, overlapping, non-randomized subsets; no human labels \\
Fact-checker CoT & 817 item indices and original scores & No recovered answer texts; cannot re-judge these historical outputs \\
HaluEval & 500 greedy, temperature and adapter answers & Prompting aggregates use smaller subsets; unmatched deltas are omitted \\
Model-size and prompt studies & Mixed logs and aggregates & Some sample sizes and judge provenance are unrecorded \\
\bottomrule
\end{tabular}
\end{table}

The historical CoT output search covered the experiment artifact tree, its March backup, and a shallow project-name search on the two archive disks. It did not recover the answer texts; this is a bounded search result, not proof that no other copy exists. Re-generating answers would constitute a new experiment rather than a re-judgment of the historical run.

\end{document}